\documentclass[letterpaper, 10 pt, conference]{ieeeconf}  

\IEEEoverridecommandlockouts                              
\usepackage{graphics} 
\usepackage{epsfig} 
\usepackage{amsmath} 
\usepackage{amssymb}  

\title{\LARGE \bf
Robot Swarming over the internet
}

\author{Will Ferenc, Hannah Kastein, Lauren Lieu, Ryan Wilson, Yuan Rick Huang,\\ J\'er\^ome Gilles, Andrea Bertozzi, Balaji R. Sharma, Baisravan HomChaudhuri,\\ Subramanian Ramakrishnan and Manish Kumar 
\thanks{Ryan Wilson and J\'er\^ome Gilles and Andrea Bertozzi are with the Department of Mathematics, University of California Los Angeles, 520 Portola Plaza Los Angeles, CA 90095-1555, U.S.A.
        {\tt\small jegilles@math.ucla.edu}}%
\thanks{Will Ferenc and Hannah Kastein and Lauren Lieu are with the Department of Engineering, Harvey Mudd College, 301 Platt Boulevard, Claremont, CA 91711, U.S.A. {\tt\small Will\_Ferenc@hmc.edu}}%
\thanks{Yuan Rick Huang is with Anteros Labs, Inc, Torrance, CA 90505, U.S.A. {\tt\small rhuang@anteroslab.com}}%
\thanks{Balaji R. Sharma, Baisravan HomChaudhuri, Subramanian Ramakrishnan and Manish Kumar are with the School of Dynamical Systems, 629 Rhodes Hall, University of Cincinnati, Cincinnati, OH 45221, U.S.A. {\tt\small sharmabr@mail.uc.edu, s.ramakrishnan@uc.edu, kumarmu@ucmail.uc.edu}}%
}

\begin{document}

\maketitle
\thispagestyle{empty}
\pagestyle{empty}

\begin{abstract}

This paper considers cooperative control of robots involving two different testbed systems in remote locations with communication on the internet. This provides us the capability to exchange robots status like positions, velocities and directions needed for the swarming algorithm. The results show that all robots properly follow some leader defined one of the testbeds. Measurement of data exchange rates show no loss of packets, and average transfer delays stay within tolerance limits for practical applications. In our knowledge, the novelty of this paper concerns this kind of control over a large network like internet.

\end{abstract}

\section{INTRODUCTION}

\begin{figure}[!t]
\includegraphics[width=\columnwidth]{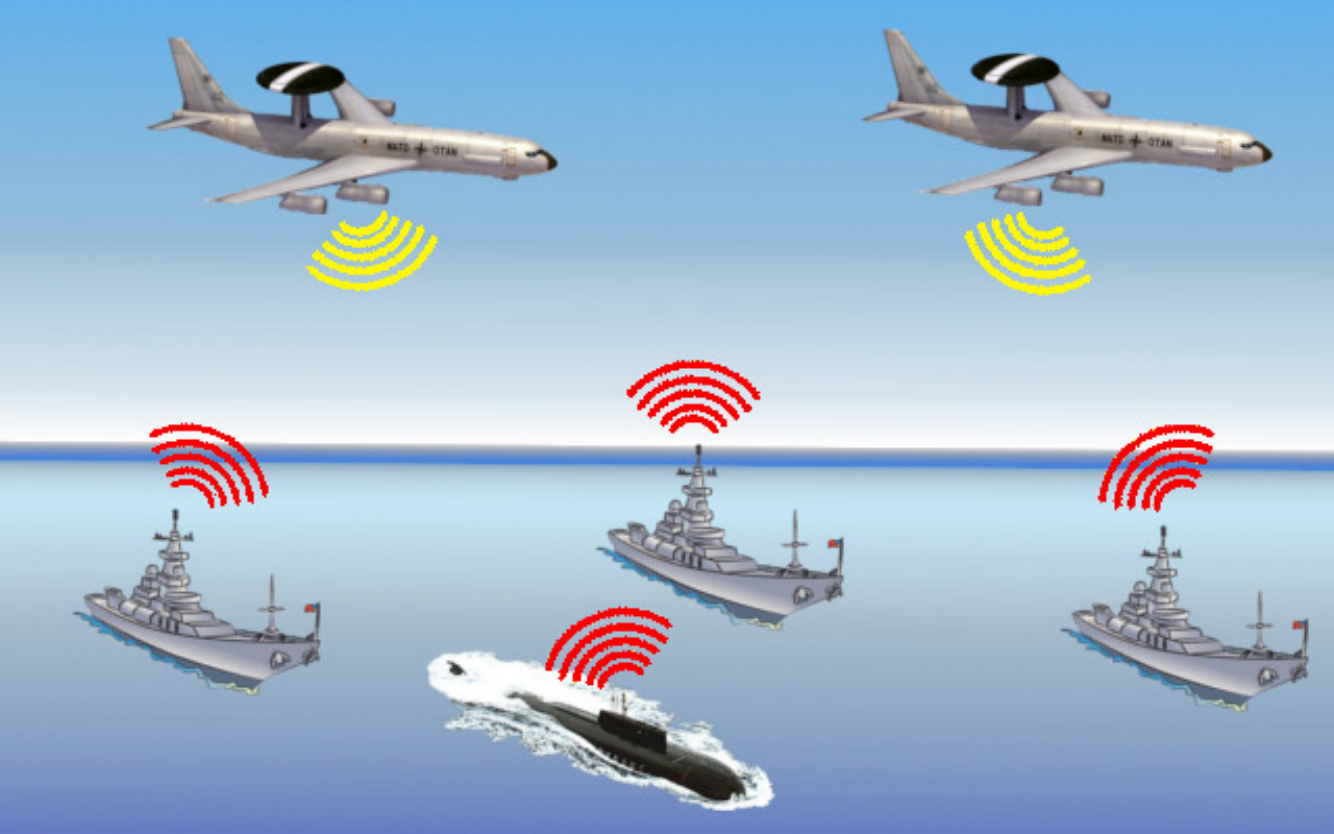}
\caption{Concept of cooperative unmaned vehicles.}
\label{fig:battlefield}
\end{figure}

The efficient co-operation between multiple agents situated at distinct locations while pursuing common objectives is an important aspect of multi-agent systems. While the topic raises fundamental questions related to a variety of fields such as communication systems and distributed co-operative control, it is of immense practical interest as well. For example, as schematically represented in Fig.~\ref{fig:battlefield}, collaborating unmanned vehicles can be used in combat or hazardous environment zones to reduce the risks for human lives. An important kind of cooperation is when agents such as robots are communicating through some network like the Internet. In this article we report on our efforts to develop and implement algorithms designed to manipulate two sets of mobile robots, situated in two geographically distinct locations, in order to cooperatively realize a common task with information exchange through the World Wide Web. While there have been many laboratory testbeds built to explore swarm algorithms, they all use local communications \cite{Li}. Our situation is novel because we use internet, with its inherent delays, as the support of our communications.\\

In order to experimentally explore algorithms for robotic applications such as the one described above, the University of California Los Angeles (UCLA) Applied Mathematics Department and the University of Cincinnati (UC) School of Dynamic Systems have built their own laboratory testbeds \cite{Gonzalez2011,2010reureport}. The testbeds allow the study of single and multiple robots tasked with missions such as path planning, target searching and environmental exploration. The primary objective of the work reported in this article was to establish a connection through the Internet between the two testbeds aimed at exploring cooperative algorithms. It was determined that a robot swarming algorithm would be an ideal candidate to test the different aspects of the communication and co-operative control mechanisms. The experiments reported in this article may be summarized as follows. In one of the testbeds, one robot was designated as the leader and was programmed to follow a predefined path. On both testbeds, the remaining robots were asked to follow the leader based upon a swarming algorithm. Each vehicle (leader and followers) broadcast their positions, speeds and headings to the others through both local radio features and the Internet. Our purpose was to test different configurations: pure simulations between the computers located at UCLA and UC, and a combination of simulations and actual robots, as steps towards meeting the larger objective of implementation of the cooperative control algorithm on actual robots on each testbed communicating over the internet.\\

The remainder of the paper is set as follows: in section \ref{sec:testbed} we give an overview of each testbed and describe how network communications are set up. In section \ref{sec:swarmalgo}, we describe the swarming algorithm used in the experiments. Section \ref{sec:expe} presents representative results from the simulations, and Section \ref{sec:conc} presents the conclusions and outlines future research directions.

\section{TESTBEDS CONFIGURATIONS AND COMMUNICATIONS}\label{sec:testbed}
\subsection{UCLA Testbed}
\begin{figure}[!h]
\includegraphics[width=\columnwidth]{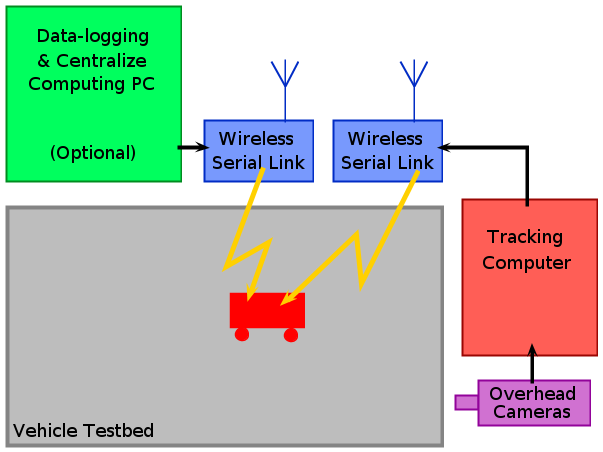}
\caption{UCLA testbed configuration.}
\label{uclatestbed}
\end{figure}

\begin{figure}[!h]
\includegraphics[width=\columnwidth]{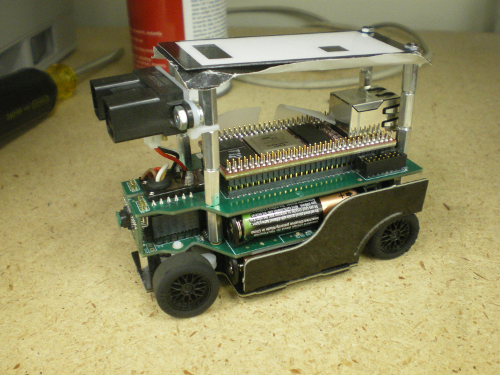}
\caption{UCLA micro-car vehicle. Its size is $50.8mm\times 101.6mm\times 45mm$.}
\label{fig:robot}
\end{figure}

The UCLA testbed is currently in its third generation and is composed of three main subsystems: (1) two overhead-cameras associated with a PC tracking system which detects each robot's position, speed and orientation (it mimicks the functionality of a GPS unit), (2) a remote terminal PC which can communicate with the robots and (3) several micro-car robotic vehicles (see Fig.~\ref{uclatestbed} for a schematic of the test bed and Fig.~\ref{fig:robot} for a typical micro-car robotic vehicle). Physically, the testbed is a 1.5m $\times$ 2.0m rectangular area. Each car carries its own identification tag on its top surface in order to be recognized by the tracking system, and is equiped with two serial radio modules (one receives the ``GPS'' information and the other is used to communicate with the remote terminal). Onboard each vehicle are two electronic boards. The lower board has all features to drive the motor, the steering servo and receive GPS data. The upper board has a Virtex4 FX-2 FPGA which is used to program the different algorithms that are being tested. More details about vehicles and a complete description of the testbed can be found in \cite{Gonzalez2011,2010reureport}.

\subsection{UC Testbed}
\begin{figure}[!h]
\includegraphics[width=\columnwidth]{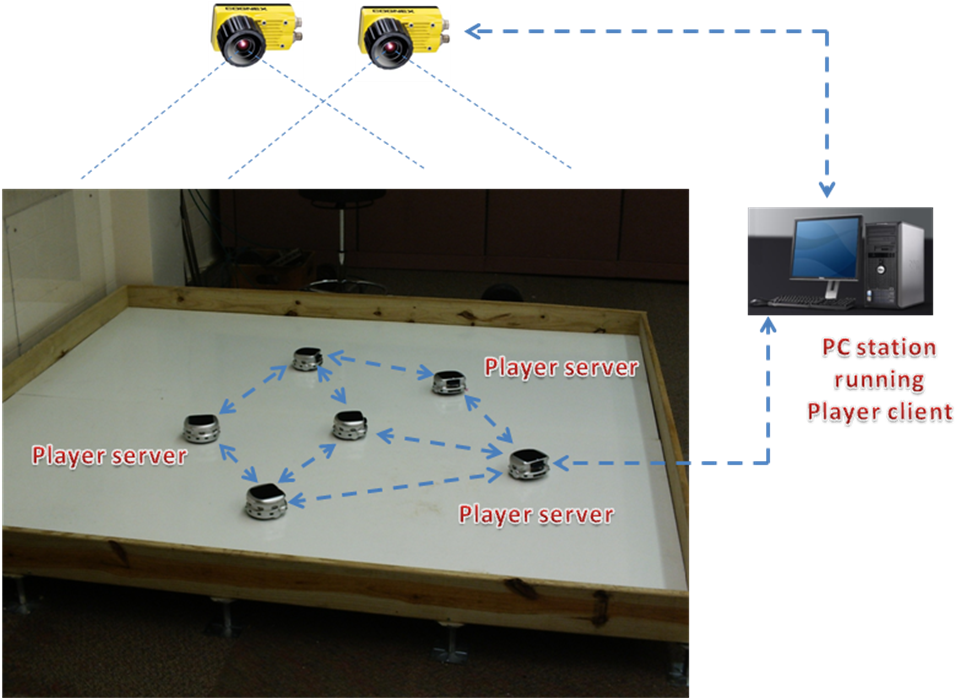}
\caption{UC testbed configuration. A Khepera III robot has a size of $130mm$ diameter and a height of $70mm$.}
\label{fig:uctestbed}
\end{figure}

The UC testbed is based on Khepera III robots \cite{khepera,devkhepera} which embeded an XScale processor, infrared and ultrasonic sensors and wireless communication capabilities. The opensource Player/Stage software is used in order to communicate with and control the robots \cite{opensource}. The Player server is installed on each Khepera robot and provides access to each sensors via TCP/IP protocol to interact with the robots. Like the UCLA's testbed, an overhead camera is connected to a tracking system installed on a remote terminal PC which mimicks the GPS function and relays global position information to the robots. This computer can also be used to exchange data with the robots by the previously mentioned TCP/IP protocol. An illustration of UC's testbed is depicted in Fig.~\ref{fig:uctestbed}

\subsection{Communication between testbeds}
\begin{figure}[!h]
\includegraphics[width=\columnwidth]{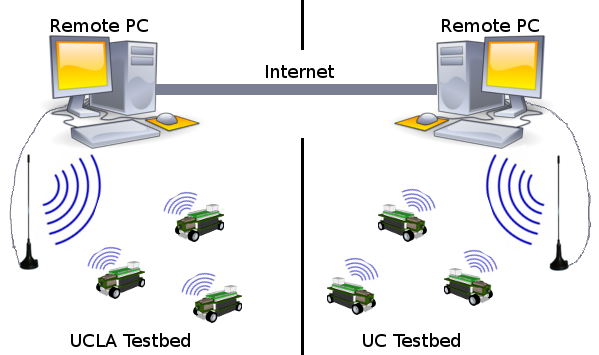}
\caption{Communication configuration between testbeds.}
\label{testbedcom}
\end{figure}

We first note that in \cite{sugisaka}, the authors discuss the use of TCP/IP communication between robots and a user interface while \cite{Turan} provides details about remote simulation using internet. Based on these papers, we determined that two different types of communication are needed for our purposes, viz. local and nonlocal. Local communications are supported by radio/wireless functions embedded in each robot. This allows each vehicle to broadcast the necessary informations (position, speed and orientation) to the others and to the remote terminal PC (RTPC). It also permits RTPC to send informations to the robots and eventually send instructions to the leader.\\
Network communications are established between each testbed RTPC located in each university by using the TCP/IP protocol over a VPN (Virtual Private Network) through the Internet. See Fig.~\ref{testbedcom} for a complete configuration view. 
The testbeds can thus communicate with each other over the established network. In addition, each robot belonging to each testbed can be updated on the status of the others (including the leader) using this link.\\
For practical reasons, we chose to use the Instrument Control Toolbox in Matlab for establishing communication. The R2011a release of Matlab allows for Matlab to act as a server, a feature hitherto inexistant. This toolbox permits us both to drive the wireless communication between robots and RTPC, and to establish the TCP/IP connection through the Internet between the two RTPCs.

We conclude this section by observing that each swarming code has its own matrix containing the information regarding position, orientation and speed for all the robots (both at UCLA and UC) under consideration. One testbed then sends an updated version of the information about its robots to the other and also listens for the updated data from the other robots. Specifically, if $X_i,Y_i,V_{x_i},V_{y_i}$ represents the two dimensional Cartesian co-ordinates of $i-th$ robot and its velocity components respectively, each message sent by a testbed has the following format:

\begin{equation}
(X_1\; X_2\; X_{N_p}\; Y_1\; Y_2\; Y_{N_p}\; V_{x_1}\; V_{x_2}\; V_{x_{N_p}}\; V_{y_1}\; V_{y_2} \; V_{y_{N_p}})
\end{equation}

where $p\in\{UCLA,UC\}$ and $N_p$ is the number of robots evolving on the corresponding testbed.\\
\section{SWARMING ALGORITHM}\label{sec:swarmalgo}
The goal of the swarming algorithm considered in this article is for a group of robots to follow a leader while avoiding collisions and maintaining formation. We note that in \cite{Moeslinger} a swarming algorithm without communication is proposed while in \cite{Scott} simulation results based on an interesting potential model are reported.  Another work \cite{Cao} uses three-dimensional mapping and graph theory to design swarming algorithms. In \cite{Lurkin}, the question of controlling large swarms of robots is addressed. After a survey of the literature keeping in mind the fact that only limited computing resources are available on robots, we chose the algorithm developed in \cite{Liu2010} where an easily implementable \cite{Nguyen}, exponential potential model is discussed. The model is qualitatively similar to the well known Morse potential that arises in molecular physics. Precisely, the following differential equations govern the dynamics of each robot except the leader which operates independently of the swarm. Practically, the velocity $v_i$ and the movement direction (resulting from $x_i$ coordinates) are provided to the motion controler of the robot.
\begin{align}
\frac{dx_i}{dt}&=v_i\\
\frac{dv_i}{dt}&=\left(\alpha-\beta\|v_i\|^2\right)v_i-\nabla U(x_i)+\sum_{j=1}^NC_0(v_j-v_i)\\
U(x_i)&=\frac{1}{2}C_l(x_i-y)^2+\sum_{j=1}^N\left(C_re^{\frac	{\|x_i-x_j\|}{l_r}}-C_ae^{\frac{\|x_i-x_j\|}{l_a}}\right)
\end{align}

In the above system of equations, $U$ is the potential function, $N$ is the number of robots on the testbed (including both the UCLA robots and the UC robots) and $y$ is the position of the leader robot. The constants associated with the model are $m, C_0 , C_l , C_r , C_a , l_r$ and $l_a$. Here $m$ refers to the robot mass, $C_0$ is the velocity alignment coefficient which dictates that the robots have identical velocities or otherwise, $C_l$ is the leader potential coefficient and represents the strength of attraction the followers experience towards the leader, $C_a$ and $C_r$ are respectively the robot attraction and repulsion coefficients that ensure collision avoidance while keeping a compact formation and $l_a$ and $l_r$ are the robot attraction and repulsion lengths, respectively. In all the experiments reported in this article, the constant values are set to $C_0=0.01, C_l=0.005,C_r=9, C_a=5, l_r=2, l_a=12, \alpha=0$ and $\beta=0.2$. The reader is referred to \cite{Liu2010} for a more detailed analytical evaluation of this control model.

\section{EXPERIMENTS}\label{sec:expe}
\subsection{Pure simulations}
\begin{figure}[!h]
\includegraphics[width=\columnwidth]{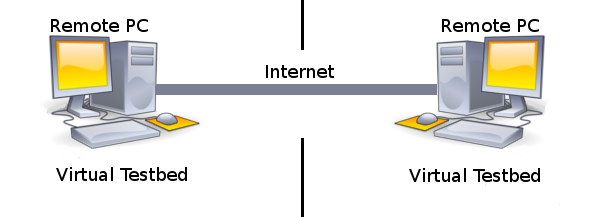}
\caption{Virtual testbeds experiment.}
\label{fig:virvir}
\end{figure}

\begin{figure}[!h]
\begin{tabular}{cc}
\includegraphics[width=0.47\columnwidth]{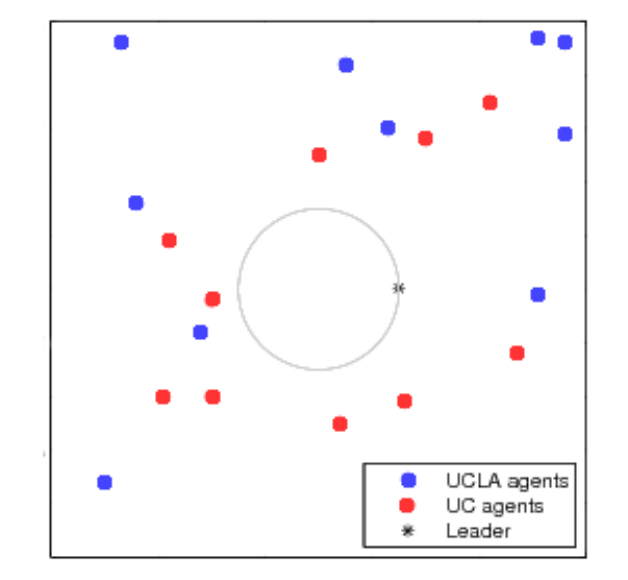}
\includegraphics[width=0.47\columnwidth]{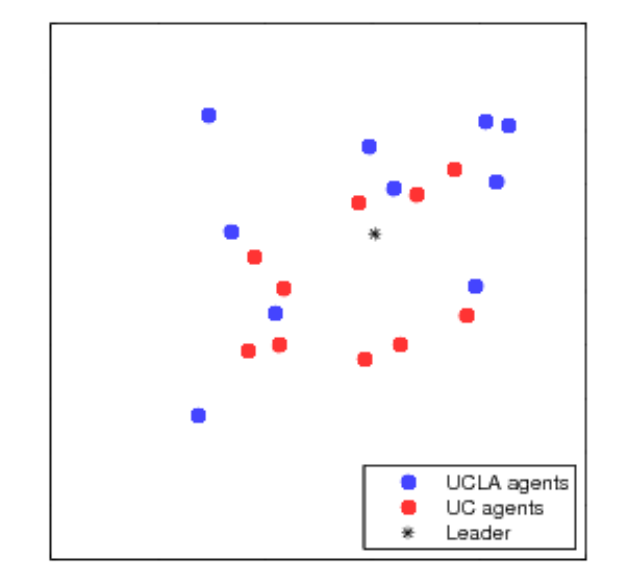}\\
\includegraphics[width=0.47\columnwidth]{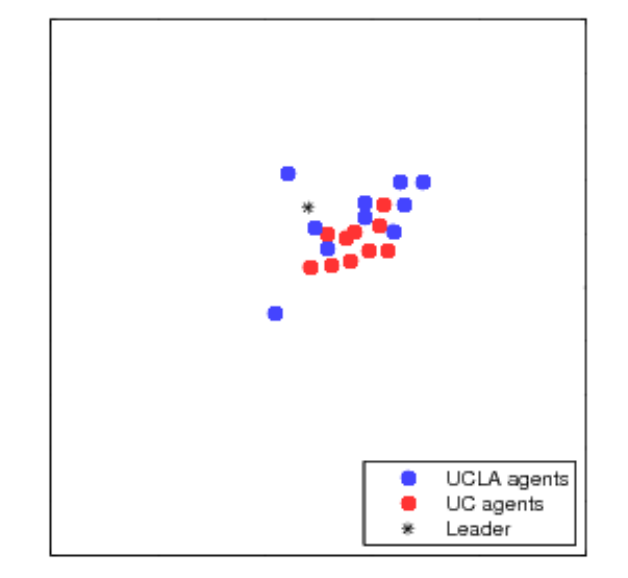}
\includegraphics[width=0.47\columnwidth]{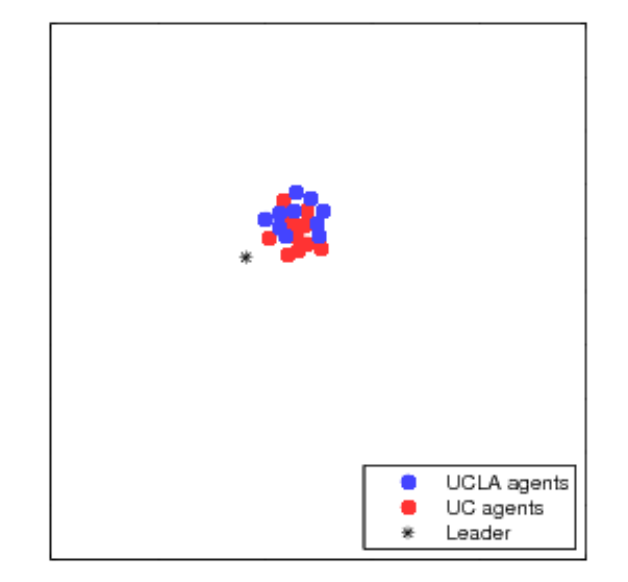}\\
\includegraphics[width=0.47\columnwidth]{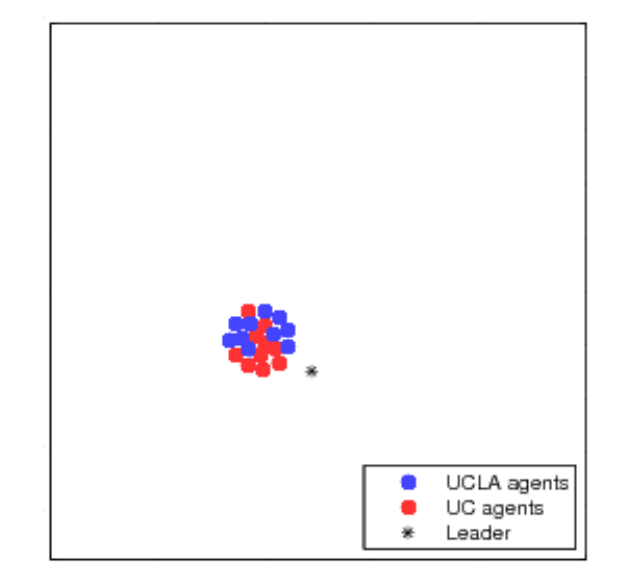}
\includegraphics[width=0.47\columnwidth]{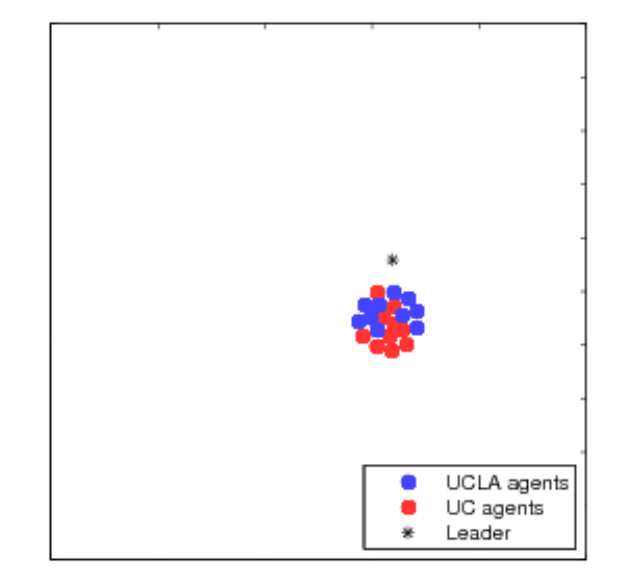}
\end{tabular}
\caption{Complete virtual swarming. The top-left frame shows the initial state of the experiment where the leader will move on the indicated circular trajectory. UCLA and UC agents are represented in blue and red respectively. Subsequent frames show the evolution of the robots during the experiment at different times.}
\label{fig:virswarm1}
\end{figure}

\begin{figure}[!h]
\includegraphics[width=\columnwidth]{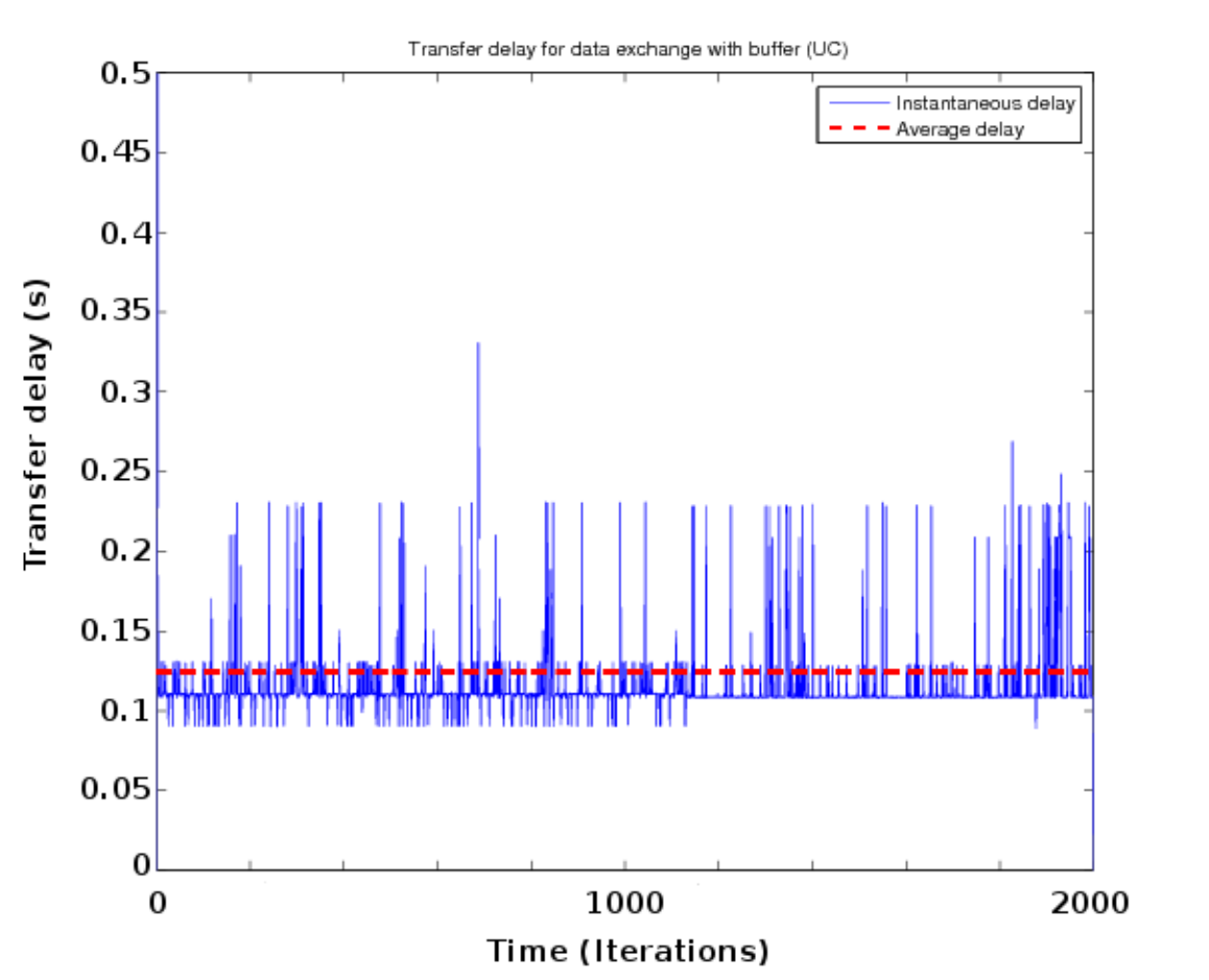}\\
\caption{Transfer delay for data exchange with buffer (UC)}
\label{fig:trdelayuc}
\end{figure}

\begin{figure}[!h]
\includegraphics[width=\columnwidth]{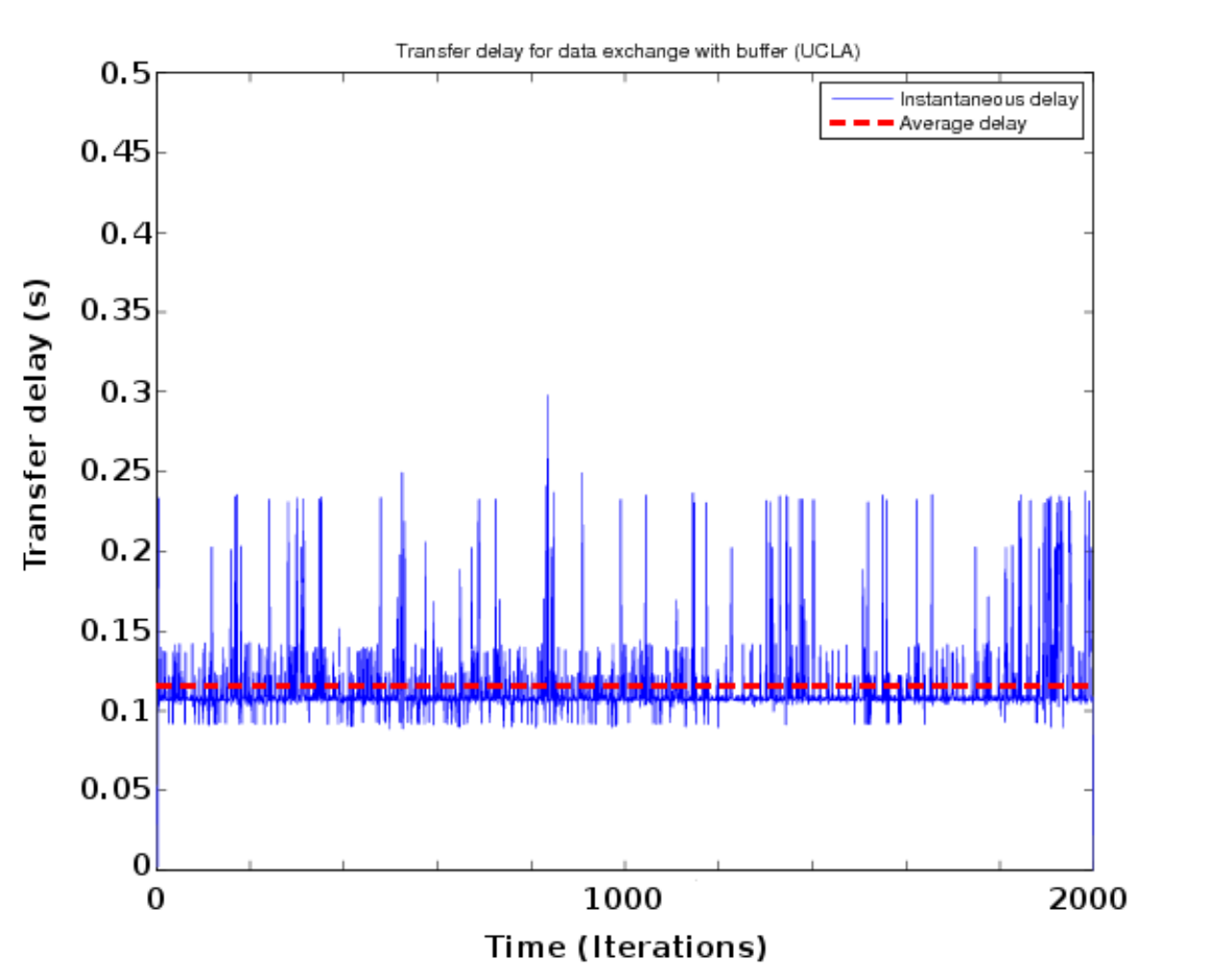}\\
\caption{Transfer delay for data exchange with buffer (UCLA)}
\label{fig:trdelayucla}
\end{figure}

\begin{figure}[!h]
\includegraphics[width=\columnwidth]{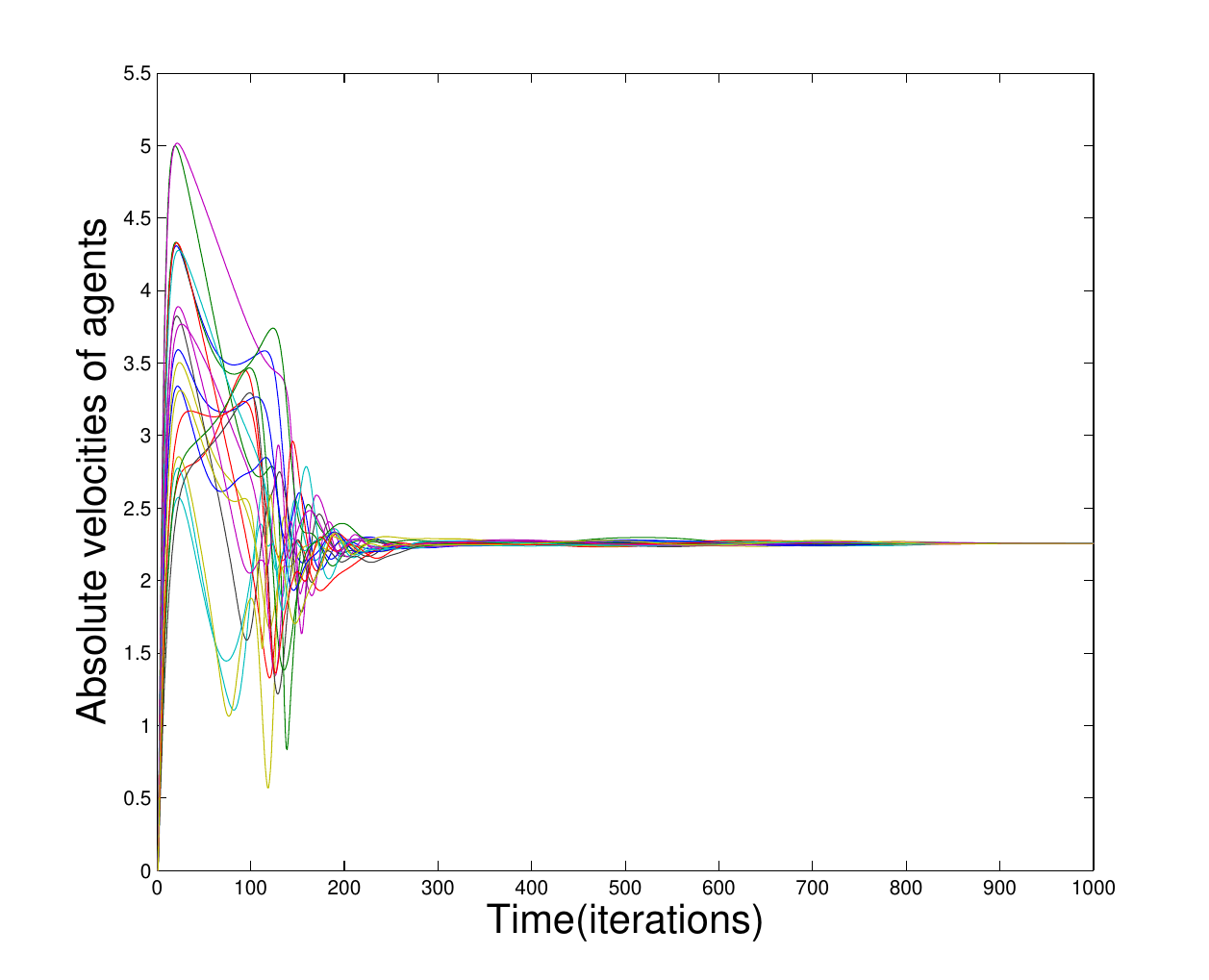}\\
\caption{Convergence of individual agent velocities to a common magnitude}
\label{fig:velocities}
\end{figure}

Our first set of experiments were based exclusively on simulations. In order to check the consistency of the swarming code (without network communications), we assumed that the leader and all the followers were on the same virtual testbed. We then simulated a variety of scenarios with different pre-programmed paths for the leader for optimal selection of gain values for the control algorithm being implemented.\\
Next, we set up the network connection between UCLA and UC, with the UC RTPC acting as the server and the UCLA RTPC as the client. A leader was defined on UCLA's side and ten virtual robots were assigned to each virtual testbed (see Fig.~\ref{fig:virvir}). For the sake of simplicity, the leader is programmed to follow a circular path in the virtual testbed. A TCP/IP socket is created connecting the RTPCs at the two ends, with I/O buffers facilitating the data exchange on a First-In-First-Out (FIFO) basis. The swarming code is implemented both sides, governing the dynamics of the virtual robots in such a manner as to follow the leader. The subfigure on top-left of Fig.~\ref{fig:virswarm1} shows the initial state at the beginning of the experiment, where all robots are initiated in a random non-overlapping configuration. The black cross represents the leader, the small blue circles UCLA's robots and red ones UC's. The gray circle in the first image indicates the path of the leader in the testing area. Intermediate states are presented in the other subfigures of Fig.~\ref{fig:virswarm1} where all robots are following the leader along its trajectory.\\
It can be observed that starting from random initial positions in the simulation area, the agents converge to a stable well-knit swarm while simultaneously in pursuit of the leader. Data exchanges were consistently fast across each control iterations, with delays observed to be within tolerance limits. Instantaneous transfer delays for the data exchange with the buffers for the RTPCs at UC and UCLA, over the length of the simulation, can be seen in Fig.~\ref{fig:trdelayuc} and Fig.~\ref{fig:trdelayucla} respectively. Average transfer delays across the length of the simulation were observed to be 0.1240 and 0.1154 seconds at the UC and UCLA ends respectively. The data exchange was synchronous with no loss of data packets during the exchange. The convergence of the magnitudes of individual agent velocities can be seen from Fig. ~\ref{fig:velocities}.

This experiment allowed us to check that data exchange through the Internet worked as expected and provided an estimate of delays expected in the data exchange. This is crucial to the implementation of the control algorithms and determination of optimal actuation rates on physical robots. The experiment allows for a strong understanding of the capabilities and limitations of data exchange over IP networks for real-time applications such as is presented in this work.

\subsection{Mixed simulation and real robots}
\begin{figure}[!h]
\includegraphics[width=\columnwidth]{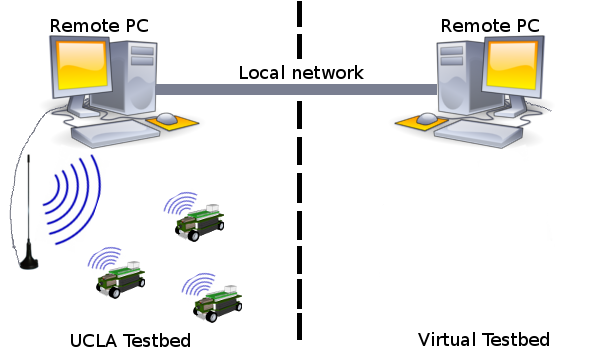}
\caption{Real robots and virtual testbed experiment.}
\label{fig:virreal}
\end{figure}

The goal of the second experiment was to check that consistency in exchanging information through a (local) network and radio communications could be achieved in the context of our application. To this end we simulated a remote RTPC on a second computer plugged on the local network which interacted with the real testbed's RTPC (see Fig.~\ref{fig:virreal}). This experiment allowed us to observe the influence of different latent times inherent to radio communications and network transmissions and investigate time-lag issues. We checked that for a reasonable number of robots, the latent times were not big constraints.\\


\section{CONCLUSIONS AND FUTURE WORKS}\label{sec:conc}

In this article, we investigated the possibility of using networks such as the Internet with their own communication constraints (time-lag in transmissions, imposed protocols, etc) in order to control the motion of groups of robots which collaborate to achieve common tasks while physically being in distinct geographical locations. We established a connection between the Applied Math Laboratory of University of California at Los Angeles and the Cooperative Distributed Systems Laboratory of the University of Cincinnati over the Internet. This connection allowed us run a swarming algorithm on both sides where only information on positions, speeds and orientations were exchanged. Results obtained from the different experimental configurations demonstrate that the concept of using the Internet as a link between robots is valid and has immense potential for a variety of applications. In practice, indeed some limitations are expected because of communication latencies through a network like the World Wide Web. Moreover, those limitations could clearly impact the number of robots involved in potential applications.

Work on implementation of the swarming algorithm across the IP network (demonstrated in simulation here), on physical robots is in progress. While simulations permit certain assumptions on the similarity in physical capabilities and actuation rates of the ground robots involved in the experiment, such assumptions are not always valid for physical robots, and the network exchanges would have to be adapted around such constraints. In future work, more complex algorithms involving cooperative behavior through the Internet such as target searching could be tested. The influence of heterogeneous control gains in the interacting swarms to offset physical constraints on the robots could be investigated too. Another interesting experiment would be to consider that the two testbeds represent the same area but at different scales. In this case, one will have two categories of robots: the ones who have a global view of the situtation and local ones which have access to more details and could broadcast finer aspects of information. Applications are also possible in remote coordination among robots in geographically separated manufacturing units for networked production and conveyance. With growing internet coverage even in remote regions, large scale terrain exploration endeavors could also benefit from such networked coordination, with local swarms at various regions of the unexplored terrain interacting over the internet for more efficient mapping at a global level.

\section{ACKNOWLEDGMENTS}

The authors gratefully acknowledge people from both universities computing resources offices to their help for establishing the different network configurations. The authors at the University of Cincinnati would like to thank Timothy Arnett, Kevin McClellan and Kaitlyn Fleming for their contributions to this work. This work is supported by the following grants: NSF-DMS-1045536, NSF-EFRI-1024765, NSF-EFRI-1024608 and NSF-DMS-0914856.


\bibliographystyle{IEEEtran}
\bibliography{biblio}

\end{document}